\documentclass[nohyperref]{article}

\usepackage{microtype}
\usepackage{graphicx}
\usepackage{subcaption}
\usepackage{booktabs} 
\usepackage{cancel}

\usepackage{hyperref}

\usepackage[accepted]{icml2022}

\usepackage{amsmath}
\usepackage{amssymb}
\usepackage{mathtools}
\usepackage{amsthm}

\usepackage{amsfonts}
\usepackage{color}
\usepackage{pifont}
\usepackage{bbm}

\usepackage[capitalize,noabbrev]{cleveref}

\theoremstyle{plain}
\newtheorem{theorem}{Theorem}[section]

\newtheorem{lemma}[theorem]{Lemma}
\newtheorem{corollary}[theorem]{Corollary}

\theoremstyle{definition}

\newtheorem{assumption}[theorem]{Assumption}

\usepackage[textsize=tiny]{todonotes}

\newcommand{\calig}[1]{\mathcal{#1}}
\newcommand{\bb}[1]{\mathbb{#1}}
\newcommand{\brac}[1]{\left(#1\right)}
\newcommand{\sbrac}[1]{\left[#1\right]}
\newcommand{\cbrac}[1]{\left\{#1\right\}}

\newcommand{\norm}[1]{\left\Vert#1\right\Vert}
\newcommand{\proj}[2]{\Pi_{#1}(#2)}
\newcommand{\taumix}{\tau_{\mathrm{mix}}}
\newcommand{\dmix}[1]{\textup{d}_\textup{mix}(#1)}
\newcommand{\bbE}{\bb{E}}
\newcommand{\distributed}[2]{#1\sim #2}
\newcommand{\tv}[2]{D_{\textup{TV}}(#1, #2)}
\def\naturals{{\mathbb N}}
\def\reals{{\mathbb R}}
\DeclarePairedDelimiter\abs{\lvert}{\rvert}
\newcommand{\Ocal}{\calig{O}}
\newcommand{\Ocaltilde}{\tilde{\calig{O}}}
\newcommand{\MLMC}{\textup{MLMC}}
\newcommand{\qedwhite}{\hfill \ensuremath{\Box}}

\newcommand{\df}{\nabla f}
\newcommand{\dF}{\nabla F}
\newcommand{\A}{\calig{A}}
\newcommand{\Scal}{\calig{S}}
\newcommand{\Pcal}{\calig{P}}
\newcommand{\Rcal}{\calig{R}}
\newcommand{\F}{\calig{F}}
\newcommand{\K}{\calig{K}}
\newcommand{\Z}{\calig{Z}}
\newcommand{\E}{\calig{E}}
\newcommand{\xmark}{\text{\ding{55}}}
\newcommand{\prob}{\bb{P}}
\DeclarePairedDelimiter\floor{\lfloor}{\rfloor}
\DeclarePairedDelimiter\ceil{\lceil}{\rceil}
\newcommand{\acronym}{MAG }

\icmltitlerunning{Adapting to Mixing Time in Stochastic Optimization with Markovian Data}

\begin{document}

\twocolumn[
\icmltitle{Adapting to Mixing Time in Stochastic Optimization with Markovian Data}





\begin{icmlauthorlist}
\icmlauthor{Ron Dorfman}{tech}
\icmlauthor{Kfir Y. Levy}{tech,Vit}
\end{icmlauthorlist}

\icmlaffiliation{tech}{Viterby Faculty of Electrical and Computer Engineering, Technion, Haifa, Israel}
\icmlaffiliation{Vit}{A Viterbi Fellow}

\icmlcorrespondingauthor{Ron Dorfman}{rdorfman@campus.technion.ac.il}

\icmlkeywords{Machine Learning, ICML}

\vskip 0.3in
]



\printAffiliationsAndNotice{}  

\begin{abstract}
We consider stochastic optimization problems where data is drawn from a Markov chain. Existing methods for this setting crucially rely on knowing the mixing time of the chain, which in real-world applications is usually unknown. We propose the first optimization method that does not require the knowledge of the mixing time, yet obtains the optimal asymptotic convergence rate when applied to convex problems. We further show that our approach can be extended to: \textbf{(i)} finding stationary points in non-convex optimization with Markovian data, and \textbf{(ii)} obtaining better dependence on the mixing time in temporal difference (TD) learning; in both cases, our method is completely oblivious to the mixing time. Our method relies on a novel combination of multi-level Monte Carlo (MLMC) gradient estimation together with an adaptive learning method.
\end{abstract}

\section{Introduction}
Stochastic optimization is one of the most important tools in machine learning (ML). A stochastic optimization problem is generally formalized as 
\begin{equation}\label{eq:so_problem}
    \min_{w\in\K}{F(w)\coloneqq\bbE_{\distributed{z}{\mu}}\sbrac{f(w;z)}}\; ,
\end{equation}
where $\K\subseteq\reals^d$ is the optimization domain, $\cbrac{w\mapsto f(\cdot;z)}$ is a set of loss functions, and $\mu$ is some unknown data distribution over $z\in\Omega$. A typical assumption in ML is that data samples are \emph{independent and identically distributed (i.i.d.)} according to $\mu$. Under this assumption, there is a wide variety of methods for solving the problem \eqref{eq:so_problem}, especially in the convex case (e.g., \citealp{nemirovski2009robust}). However, this setting fails to capture many real-world problems in economics, control, and reinforcement learning (RL), in which samples are correlated~\cite{hamilton1994autoregressive, kushner2003stochastic, sutton2018reinforcement}.


A natural and popular way to characterize temporal data dependencies is to assume that the data evolves according to a \emph{Markov chain}~\cite{levin2017markov}.  This gives rise to  stochastic optimization problems with \emph{Markovian data}: In such problems, the objective is the same as the one in Eq.~\eqref{eq:so_problem}; nevertheless, the learner cannot access i.i.d. samples from $\mu$, but rather a sequence $z_1,z_2,\ldots$ of \emph{dependent} samples that arrive from a Markov chain whose stationary distribution is $\mu$. As we next describe, solving such problems poses a challenge to standard training methods.


A key parameter in the analysis of optimization problems with Markovian data is the \textit{mixing time} $\taumix$, which is the minimal time difference between two approximately independent samples (see formal definition in Eq.~\eqref{eq:mix_time_def}). Devising efficient algorithms for such problems crucially rely on knowing $\taumix$, and a lack of this knowledge may lead to a significant performance degradation. 

Prior work on stochastic optimization with Markovian data assume the knowledge of the mixing time and use it explicitly to obtain favorable performance guarantees~\cite{duchi2012ergodic, bresler2020least, jain2021streaming}. Unfortunately, in real world applications, the mixing time is usually unknown in-advance, which hurts the practicality of these approaches. Alternatively, one may consider estimating the mixing time directly from data; however, the sample complexity of mixing time estimation is quite large, and it depends on specific properties of the Markov chain~\cite{hsu2017mixing, wolfer2019estimating, wolfer2020mixing}. 

\begin{table*}[t]
    \caption{Comparison of methods for stochastic optimization with Markovian data.}
    \begin{center}
        \begin{tabular}{*{5}{c}}
            \toprule
            {\centering\textbf{Method}}  & \multicolumn{1}{c} {\centering\textbf{Learning}} & \multicolumn{1}{c} {\centering\textbf{Oblivious}} & \multicolumn{1}{c} {\textbf{\# observed samples}} & \multicolumn{1}{c} {\centering\textbf{Convergence}} \\
            & \multicolumn{1}{c} {\centering\textbf{rate}} & \multicolumn{1}{c} {\centering\textbf{to $\taumix$}} & \multicolumn{1}{c} {\centering\textbf{(expectation)}} & \multicolumn{1}{c} {\centering\textbf{rate}} \\
            \midrule
            SGD & $1/\sqrt{T}$ & \checkmark & $T$ & $\Ocal\brac{\taumix/\sqrt{T}}$ \\
            EMD~\citep{duchi2012ergodic} & $1/\sqrt{\taumix T}$ & $\xmark$ & $T$ &  $\Ocal\brac{\sqrt{\taumix/T}}$ \\
            SGD-DD & $1/\sqrt{T}$ & $\xmark$ & $T$ & $\Ocal\brac{\sqrt{\taumix/T}}$ \\
            MAG (This work) & AdaGrad~\eqref{eq:adagrad_lr} & \checkmark & $\Ocaltilde(T)$ & $\Ocaltilde\brac{\sqrt{\taumix/T}}$ \\
            \bottomrule
        \end{tabular}
    \end{center}
    \label{tab:comparison_props}
\end{table*}

In this work, we propose an efficient first-order algorithm for stochastic optimization with Markovian data that does not require the knowledge of the mixing time, yet obtains convergence rates as if the mixing time was known. In that sense, our method is \textit{adaptive} to the mixing time. Focusing on the stochastic convex optimization setting, we show that after $T$ iterations and $\Ocaltilde(T)$ data samples, our method obtains a convergence rate of $\Ocaltilde(\sqrt{\taumix/T})$, which is optimal up to logarithmic factors~\cite{duchi2012ergodic}.

Our contribution goes beyond the convex setting as we also provide convergence results for two additional settings: \textbf{(i)} in the non-convex case, 
we show that after observing $\Ocaltilde(T)$ samples, our method finds an $\epsilon$-stationary point, where $\epsilon=\Ocaltilde((\taumix/T)^{1/4})$, for any smooth and bounded function, 
and \textbf{(ii)} we study temporal difference (TD) learning with linear function approximation~\cite{tsitsiklis1997analysis}, which is widely used in RL for value function estimation, and provide a finite-time bound that is $\sqrt{\taumix}$ times better than the bound recently achieved by \citet{bhandari2018finite}.
Finally, we validate the performance of our algorithm on a synthetic linear regression task.

From a technical point of view, our results stem from a novel new first-order method that combines two ideas: estimating the gradient using a multi-level Monte Carlo (MLMC) technique~\cite{giles2015multilevel, blanchet2015unbiased} and using the adaptive learning rate of AdaGrad~\cite{duchi2011adaptive}. We therefore term our method \textit{\textbf{M}LMC-\textbf{A}da\textbf{G}rad (MAG)}.

\subsection{Related Work}
The most popular method for solving stochastic optimization problems is stochastic gradient descent (SGD), which is defined by the following update rule:
\begin{equation*}\label{eq:sgd}
    w_{t+1} = \proj{\K}{w_t - \eta_t g_t},\quad t=1,\ldots,T\; ,
\end{equation*}
where $\eta_t$ is the learning rate at time $t$, and $g_t$ is an estimate of $\nabla F(w_t)$, e.g., $g_t=\nabla f(w_t;z_t)$.
SGD has been extensively studied in the literature~\cite{robbins1951stochastic, polyak1992acceleration, nesterov2012make}. Key ingredient in the analysis of SGD in the i.i.d. case is the conditional unbiasedness of the stochastic gradients: Since samples are independent, the randomness in $g_t$ is independent of the iterate $w_t$, and it holds that $\bbE[g_t|w_t] = \dF(w_t)$. This property enables to obtain the optimal $\Ocal(1/\sqrt{T})$ convergence rate for convex problems by choosing the learning rate proportionally to $1/\sqrt{T}$~\cite{nemirovski2009robust,agarwal2012information}.

In the non-i.i.d. case, however, the gradients are not conditionally unbiased because $w_t$ depends on the samples up to time $t-1$, which are no longer independent of $g_t$. Therefore, the non-i.i.d. setting introduces an additional bias term that needs to be properly controlled to ensure fast convergence. To that end, \citet{duchi2012ergodic} used a smaller learning rate proportional to $1/\sqrt{T\taumix}$. They analyzed Ergodic Mirror Descent (EMD), a variant of SGD that also applies to non-Euclidean spaces and showed it achieves their derived lower bound of $\Omega(\sqrt{\taumix/T})$.

Another simple technique for reducing the bias is SGD Data-Drop (SGD-DD,~\citealp{bresler2020least}), which applies a single stochastic gradient update every $\taumix$ samples and discards the rest of the data. Intuitively, SGD-DD is approximately equivalent to performing $T/\taumix$ standard SGD updates in $T$ steps, which implies the optimal $\Ocal(\sqrt{\taumix/T})$ convergence rate. Note, however, that this method strongly relies on prior knowledge of the mixing time. 

The third technique for controlling the bias term is decorrelating samples by using a buffer to store previous data. This technique is inspired by the experience replay method~\cite{lin1992self, mnih2013playing}, which is extensively used in RL to break temporal correlations, and its variants were used and theoretically analyzed in a series of recent papers~\cite{bresler2020least, jain2021streaming, agarwal2021online}. On the downside, using a buffer requires higher memory complexity, and the variants used in the aforementioned works have a `gap' parameter that encodes the mixing time. 

Our method relies on an alternative gradient estimator, which is based on the MLMC technique. MLMC estimation was previously used in ML and optimization applications, such as distributionally robust optimization~\cite{levy2020large, hu2021bias} and latent variable models~\cite{shi2021multilevel}. Our estimator is similar to that of \citet{asi2021stochastic}, which they use to estimate proximal points and gradients of the Moreau envelope. Here, we use the MLMC technique for obtaining a gradient estimator with low-bias for stochastic optimization with Markovian data; see Table~\ref{tab:comparison_props} for comparison with existing methods.

Finally, we mention the work of \citet{sun2018markov}, in which they analyze SGD with Markovian data for both convex and non-convex problems. However, they consider a different definition of mixing time than ours, and their method is not adaptive to the mixing time.
\section{Background}
We recapitulate Markov chains and mixing time and present the notations and assumptions used throughout this manuscript.
\subsection{Markov Chains and Mixing Time}
Let $z_1,z_2,\ldots$ be an ergodic time-homogenous Markov chain over a finite state space $\Omega$ with stationary~distribution~$\mu$. We define the total variation distance of two distributions $P$ and $Q$ defined over the same probability space $(\Omega,\F)$ as
\begin{equation*}
    \tv{P}{Q}\coloneqq\sup_{A\in\F}{\abs{P(A) - Q(A)}}\; .
\end{equation*}
For every $z\in\Omega$, let $P^{t}(z,\cdot)$ denote the conditional distribution over $z_{t+1}$ given that $z_1=z$. We define the following distance from stationarity measure~\citep{levin2017markov, bresler2020least}:
\begin{equation}\label{eq:d_mix}
    \dmix{t}\coloneqq \sup_{z\in\Omega}{\tv{P^t(z,\cdot)}{\mu}}\; .
\end{equation}
This measure is a non-increasing function of $t$, and it is used to define the \textit{mixing time} of the Markov chain:
\begin{equation}\label{eq:mix_time_def}
\begin{aligned}
    \taumix(\epsilon) &\coloneqq \inf\cbrac{t: \dmix{t}\leq \epsilon}, \\ \taumix &\coloneqq \taumix(1/4)\; . 
\end{aligned}
\end{equation}
The mixing time definition implies a useful result; for every $\ell\in\naturals$ it holds that
\begin{equation}\label{eq:dmix}
    \dmix{\ell\taumix}\leq 2^{-\ell}\; .
\end{equation}

\subsection{Notations and Assumptions}
Let $d$ denote the dimension of the problem. We use $\norm{\cdot}$ to denote the $L_2$ norm of vectors in $\reals^d$, and let $w^*$ denote the minimum of $F$ over $\K\subseteq\reals^d$. For every $n\in\naturals$, let $\sbrac{n} = \{1,\ldots,n\}$. $\F_t$ denotes the $\sigma$-field induced by all the Markov chain samples up to time $t$. We use $\prob_{t}$ and $\bbE_{t}$ to denote the conditional probability and expectation given $\F_t$, respectively.

Depending on the convexity of the objective, we make some of the following assumptions, which are standard when analyzing gradient-based algorithms.
\begin{assumption}[Bounded gradients]\label{assump:bounded_gradients}
    There exist a constant $G>0$ such that
    \[
        \norm{\df(w;z)}\leq G,\quad \forall w\in\K, z\in\Omega\; .
    \]
\end{assumption}
\begin{assumption}[Bounded domain]\label{assump:bounded_domain}
    The domain $\K$ is bounded with diameter $D$, i.e., $\sup_{x,y\in\K}{\norm{x-y}}\leq D$.
\end{assumption}
\begin{assumption}[Bounded function values]\label{assump:bounded_objective}
    There exist a constant $M>0$ such that
    \[
        \abs{F(w)}\leq M,\quad \forall w\in\K\; .
    \]
\end{assumption}
\begin{assumption}[Smoothness]\label{assump:smoothness}
    The objective $F$ is $L$-smooth, i.e., $F$ is differentiable and for every $x,y\in\K$ the following holds:
    \[
        \norm{\dF(x) - \dF(y)}\leq L\norm{x - y}\; .
    \]
    Equivalently, for every $x,y\in\K$ it holds that
    \[
        F(y) \leq F(x) + \dF(x)^\top(y-x) + \frac{L}{2}\norm{x-y}^2\; .
    \]
\end{assumption}

\section{MAG: Combining Multi-level Monte Carlo Gradient Estimation with AdaGrad}\label{sec:MAG}
In this section we present MAG: an algorithm for stochastic optimization with Markovian data. Recall that the objective is given in Eq.~\eqref{eq:so_problem}, and we have access to samples $z_1,z_2,\ldots$ drawn from a Markov chain.


MAG utilizes an MLMC gradient estimation technique that effectively transduces bias into variance; that is, our gradient estimator enjoys a low bias, but exhibits a high variance that depends on the mixing time (see Lemma~\ref{lem:mlmc_bounds}). To cope with this dependence, we then resort to AdaGrad~\cite{duchi2011adaptive}, which is known to implicitly adapt to the variance of the stochastic gradients~\cite{levy2018online}. We describe MAG in Algorithm~\ref{alg:mlmc_adagrad}, and state its main convergence guarantee for convex objectives in Theorem~\ref{thm:convex_convergence}.  

\paragraph{MLMC Gradient Estimator:} Before we present our suggested gradient estimator, we introduce a modified notation for the Markov chain samples, which will make the presentation and analysis more convenient. Since computing the new gradient might require more than a single Markov chain sample per-iteration, we let $N_t$ denote the (stochastic) number of samples used for computing the estimator in the $t$-th iteration (in SGD, for example, $N_t$ is deterministic and equals $1$ for all $t$). Then, we let $z_t^{(i)}$ denote the $i$-th sample in the $t$-th iteration for $i\in\sbrac{N_t}$; that is, we write the Markov chain as $z_1^{(1)},\ldots,z_1^{(N_1)},z_2^{(1)},\ldots,z_{2}^{(N_2)},\ldots$. Accordingly, we slightly modify the $\sigma$-field $\F_t$ to also include the stochastic number of samples per-iteration, such that $\F_t~=~\sigma(z_1^{(1)},\ldots,z_1^{(N_1)},\ldots,z_{t}^{(1)},\ldots,z_{t}^{(N_t)}, N_1,\ldots,N_t)$. We denote the average of stochastic gradients evaluated at $w_t$ over $2^{j}$ samples from the Markov chain by
\begin{equation}\label{eq:minimbatch_grad_est}
    g_t^{j}\coloneqq\frac{1}{2^j}\sum_{i=1}^{2^j}{\df(w_t;z_t^{(i)})}\; ,
\end{equation}
and define our MLMC estimator as follows: \\
Draw $\distributed{J_t}{\text{Geom}(\frac{1}{2})}$ and set
\begin{equation}\label{eq:mlmc_grad_estimator}
    g_t^{\MLMC} = g_t^{0} + \begin{cases}
    2^{J_t}(g_t^{J_t} - g_t^{J_t-1}), &2^{J_t}\leq T \\
    0, &\text{otherwise}
    \end{cases}\; .
\end{equation}
Note that computing this estimator requires $N_t=2^{J_t}$ samples from the Markov chain. Importantly, in each iteration, $N_t$ is independent of the previous draws $N_1,\ldots,N_{t-1}$ and of the Markovian data. 

As the following lemma implies, our estimator has the same bias as a simple average of $\Theta(T)$ noisy gradients, but it only requires $\Ocaltilde(1)$ data samples in expectation. On the downside, it has higher variance compared to the average of $T$ stochastic gradient. 
\begin{lemma}\label{lem:mlmc_bounds}
Let $j_{\max}=\max\cbrac{j\in\naturals: 2^{j}\leq T}=\floor{\log{T}}$, and let $F:\K\to\reals$ be a function with bounded gradients (Assumption~\ref{assump:bounded_gradients}). Then, 
\[
    \bbE_{t-1}\sbrac{g_t^\MLMC} = \bbE_{t-1}[g_t^{j_{\max}}],
\]
and, 
\[
    \bbE\sbrac{\norm{g_t^\MLMC}^2}\leq \Ocaltilde\brac{G^2 \taumix}\; .
\]
Moreover, the expected number of stochastic gradient evaluations required for computing $g_t^\MLMC$ is $\Ocal(\log{T})$.
\end{lemma}
\begin{proof}
Since $\prob(J_t = j) = 2^{-j}$, by explicitly writing the expectation w.r.t. $J_t$, we get a telescoping sum:
\begin{align*}
    \bbE_{t-1}\sbrac{g_t^\MLMC} &= \bbE_{t-1}\sbrac{g_t^{0}} \\ &\quad+ \sum_{j=1}^{j_{\max}}{\prob(J_t=j) 2^{j}\bbE_{t-1}[g_t^{j} - g_t^{j-1}]} \\ &= \bbE_{t-1}[g_t^{j_{\max}}]\; .
\end{align*}
Proving the second order bound is less trivial, and it requires some care. Our strategy relies on a concentration bound for estimating the gradient using the average of stochastic gradients based on Markovian samples (see Lemma~\ref{lem:concentraion_mc}). Using this result, we derive the following bound (given in Lemma~\ref{lem:expectation_and_second_order_bound}) for every $N\leq T$:
\begin{align}\label{eq:second_order_bound}
    \bbE&\sbrac{\norm{\frac{1}{N}\sum_{i=1}^{N}{\df(w_t;z_t^{(i)})} - \dF(w_t)}^2}\leq \Ocaltilde\brac{\frac{G^2\taumix}{N}}.
\end{align}
Using $\norm{x+y}^2\leq 2\norm{x}^2 + 2\norm{y}^2$ and the boundness of the gradients under Assumption~\ref{assump:bounded_gradients}, we have that $\bbE[\lVert g_t^{\MLMC}\rVert^2] \leq 2\bbE[\lVert g_t^\MLMC-g_t^{0}\rVert^2] + 2G^2$.
By explicitly writing the expectation in the right side, we obtain
\begin{align}\label{eq:second_moment_mlmc_minus_ancor}
    \bbE\sbrac{\lVert g_t^\MLMC-g_t^{0}\rVert^2} &= \sum_{j=1}^{j_{\max}}{\prob(J_t=j) 2^{2j}\bbE\sbrac{\lVert g_t^{j} - g_t^{j-1}\rVert^2}}  \nonumber \\ &= \sum_{j=1}^{j_{\max}}{2^{j}\bbE\sbrac{\lVert g_t^{j} - g_t^{j-1}\rVert^2}} \; .
\end{align} 
Since $2^{j_{\max}}\leq T$, we can bound $\bbE[\lVert g_t^{j} - g_t^{j-1}\rVert^2]$ using Eq.~\eqref{eq:second_order_bound} as follows:
\begin{align*}
    \bbE\sbrac{\lVert g_t^{j} - g_t^{j-1}\rVert^2}&\leq 2\bbE\sbrac{\lVert g_t^{j} - \dF(w_t)\rVert^2} \\ &\quad+ 2\bbE\sbrac{\lVert g_t^{j-1} - \dF(w_t)\rVert^2}\\ &\leq \Ocaltilde\brac{\frac{G^2\taumix}{2^j}}\; .
\end{align*}
Plugging-in this bound back to Eq.~\eqref{eq:second_moment_mlmc_minus_ancor}, we get that
\[
    \bbE\sbrac{\lVert g_t^\MLMC-g_t^{0}\rVert^2} \leq \Ocaltilde\brac{G^2\taumix j_{\max}} = \Ocaltilde\brac{G^2\taumix}\; ,
\]
which implies the second order bound. Finally, since the estimator requires $2^{j}$ gradient evaluations when $J_t=j$, the expected number of gradient evaluations for computing the estimator is $1+\sum_{j=1}^{j_{\max}}{\prob(J_t=j)(2^{j}-1)} = \Ocal(\log{T})$. 
\end{proof}
\paragraph{Adaptive Learning Rate:} Since the variance (or second moment) of the proposed estimator depends on the mixing time, to ensure that our method does not require the knowledge of $\taumix$, we set the learning rate according to AdaGrad, which is adaptive to the variance. Specifically, we use the scalar version of AdaGrad, also known as AdaGrad-Norm~\citep{levy2017online, ward2019adagrad}:
\begin{equation}\label{eq:adagrad_lr}
    w_{t+1} = \proj{\K}{w_t - \eta_t g_t}; \quad\eta_t = \frac{\alpha}{\sqrt{\sum_{k=1}^{t}{\norm{g_k}^2}}}\; ,
\end{equation}
for some scaling parameter $\alpha>0$. In the context of online convex optimization, this learning rate has been used for achieving a second-order regret bound~\citep{rakhlin2013optimization, levy2017online}:
\begin{lemma}[AdaGrad's regret bound]\label{lem:adagrad_regret_bound}
Let $F:\K\to\reals$ be a convex function over a bounded domain (Assumption~\ref{assump:bounded_domain}). Then, the iterates defined by AdaGrad-Norm~\eqref{eq:adagrad_lr} with~$\alpha~=~D/\sqrt{2}$ satisfy 
\[
    \sum_{t=1}^{T}{g_t^\top(w_t - w^*)}\leq D\sqrt{2\sum_{t=1}^{T}{\norm{g_t}^2}}\; .
\]        
\end{lemma}
Using this bound, we are able to obtain an implicit adaptivity to the mixing time. The following result establishes the convergence of MAG (Algorithm~\ref{alg:mlmc_adagrad}) for convex problems; see Appendix~\ref{app:proof_thm_convex_case} for a full proof.

\begin{algorithm}[t]
\caption{MAG (\textbf{M}LMC \textbf{A}da\textbf{G}rad)}
    \begin{algorithmic}\label{alg:mlmc_adagrad}
    \STATE{\bfseries Input:} Number of iterations $T$, scaling parameter $\alpha>0$. 
    Initialize $w_1\in\K$ 
    \FOR{$t=1,\ldots,T$}
        \STATE Draw $\distributed{J_t}{\text{Geom}(\frac{1}{2})}$
        \STATE Compute $g_t$ using \eqref{eq:mlmc_grad_estimator}
        \STATE Set $\eta_t = \alpha/\sqrt{\sum_{k=1}^{t}{\norm{g_k}^2}}$
        \STATE Update $w_{t+1} = \proj{\K}{w_t - \eta_t g_t}$
    \ENDFOR \\
    \textbf{return} {$\bar{w}_T = \frac{1}{T}\sum_{t=1}^{T}{w_t}$.}
\end{algorithmic}
\end{algorithm}

\begin{theorem}\label{thm:convex_convergence}
Let $F:\K\to\reals$ be a convex function with bounded gradients (Assumption~\ref{assump:bounded_gradients}) over a bounded domain (Assumption~\ref{assump:bounded_domain}). Then, the output of Algorithm~\ref{alg:mlmc_adagrad} with $\alpha=D/\sqrt{2}$ satisfies
\begin{equation*}
    \bbE[F(\bar{w}_T) - F(w^*)]\leq\Ocaltilde\brac{GD\sqrt{\frac{\taumix}{T}}}\; .
\end{equation*}
\end{theorem}
\noindent\textit{Proof Sketch.} By the convexity of $F$, we have $\bbE[F(w_t) - F(w^*)]\leq \bbE[\dF(w_t)^\top(w_t - w^*)]$. Now, adding and subtracting $g_t$ (the MLMC estimator used in Algorithm~\ref{alg:mlmc_adagrad}), we can write \begin{align*}
    \bbE[F(w_t) - F(w^*)] &\leq \bbE\sbrac{g_t^\top(w_t-w^*)} \\ &\quad+ \bbE\sbrac{(\dF(w_t) - g_t)^\top(w_t-w^*)} .
\end{align*}
By Jensen's inequality, we get  
\begin{align*}
    \vspace{-1em}
    \bbE\sbrac{F(\bar{w}_T) - F(w^*)} &\leq \frac{1}{T}\bbE\sbrac{\sum_{t=1}^{T}{g_t^\top(w_t - w^*)}} \\ &\hspace{-2em}+ \frac{1}{T}\bbE\sbrac{\sum_{t=1}^{T}{(\dF(w_t) - g_t)^\top(w_t-w^*)}}.
    \vspace{-1em}
\end{align*}
We can bound the first term by $\Ocaltilde\brac{GD\sqrt{\taumix/T}}$ using the regret bound of AdaGrad in Lemma~\ref{lem:adagrad_regret_bound} and the second order bound in Lemma~\ref{lem:mlmc_bounds}. For bounding the second term, we define $b_t(g) = (\dF(w_t) - g)^\top(w_t-w^*)$. Since $w_t$ is measurable w.r.t. $\F_{t-1}$, Lemma~\ref{lem:mlmc_bounds} implies that
\begin{align*}
    \bbE[b_t(g_t)] &= \bbE\sbrac{\bbE_{t-1}\sbrac{b_t(g_t)}} = \bbE[b_t(g_t^{j_{\max}})]\; ,
\end{align*}
where $g_t^{j}$ is defined in Eq.~\eqref{eq:minimbatch_grad_est} and $j_{\max} = \floor{\log{T}}$. From Cauchy-Schwarz inequality, we can bound $\bbE[b_t(g_t)] = \bbE[b_t(g_t^{j_{\max}})]$ by $D\cdot\bbE[\lVert g_t^{j_{\max}} - \dF(w_t)\rVert]$. Using Eq.~\eqref{eq:second_order_bound} and Jensen's inequality, we obtain the following bound: 
\begin{equation}\label{eq:norm_bound}
    \bbE[\lVert g_t^{j_{\max}} - \dF(w_t)\rVert] \leq\Ocaltilde\brac{G\sqrt{\frac{\taumix}{T}}}\; ,     
\end{equation}
which in turn implies that $\bbE[(\dF(w_t) - g_t)^\top(w_t-w^*)]\leq\Ocaltilde\brac{GD\sqrt{\taumix/T}}$. Finally, having this upper bound for each element means it also bounds the average. 
$\qedwhite$ \\

Theorem~\ref{thm:convex_convergence} implies that \acronym achieves an optimal convergence rate up to logarithmic factors~\citep{duchi2012ergodic}. Importantly, it does not require knowing $\taumix$ in advance, which makes it adaptive to the mixing time.
\section{Extension to Non-Convex Problems}\label{sec:nonconvex_MAG}
Our method can also be applied to non-convex problems with Markovian data for finding an approximate stationary point; that is, a point $w$ for which $\bbE[\lVert\dF(w)\rVert]$ is small. 

In the i.i.d.~setting, \citet{ghadimi2013stochastic}~showed~that~for any smooth objective, SGD converges to a stationary point at a rate of $\Ocal(1/T^{1/4})$ under appropriate selection of learning rate; this rate is known to be tight~\cite{arjevani2019lower}. In the context of adaptive methods, AdaGrad was shown to achieve the same rate without requiring the knowledge of the smoothness nor the noise variance~\cite{zhou2018convergence, li2019convergence, ward2019adagrad, defossez2020convergence}

In this section, we show that MAG obtains a convergence rate of $\Ocaltilde((\taumix/T)^{1/4})$ when applied to stochastic non-convex optimization with Markovian data (see Corollary~\ref{cor:nonconvex_convergence}). Our analysis assumes bounded objective; such problems include neural networks with bounded activation in the output layer (e.g., sigmoid or softmax) as well as some loss functions in robust non-convex problems~\citep{barron2019general}.


We consider unconstrained optimization and use $\Delta_{t}\coloneqq F(w_t) - \min_{w\in\reals^d}{F(w)}$ to denote the suboptimality at time $t$. The following lemma characterizes the iterates of Algorithm~\ref{alg:mlmc_adagrad} when applied to smooth functions.
\begin{lemma}\label{lem:sgd_nonconvex_lemma}
    Let $F:\reals^d\to\reals$ be an  $L$-smooth function (Assumption~\ref{assump:smoothness}). Then, the iterates of Algorithm~\ref{alg:mlmc_adagrad} satisfy
    \begin{align*}
        \sum_{t=1}^{T}{\norm{\dF(w_t)}^2} &\leq \frac{\Delta_{\max}}{\eta_T} + \frac{L}{2}\sum_{t=1}^{T}{\eta_t\norm{g_t}^2} \\ &\qquad+ \sum_{t=1}^{T}{(\dF(w_t) - g_t)^\top\dF(w_t)}\; ,
    \end{align*}
    where $\Delta_{\max} = \max_{t\in\sbrac{T}}{\Delta_t}$. 
\end{lemma}
\begin{proof}
    By the smoothness of $F$, for every $x,y\in\K$ we have $F(y)\leq F(x) + \dF(x)^\top(y-x) + \frac{L}{2}\norm{y-x}^2$. Plugging-in the update rule $w_{t+1} = w_t - \eta_t g_t$, we obtain that
    \begin{align*}
        F(w_{t+1}) &\leq F(w_t) - \eta_t \dF(w_t)^\top g_t + \frac{L\eta_t^2}{2}\norm{g_t}^2 \\ &= F(w_t) - \eta_t\norm{\dF(w_t)}^2 \\ &\quad+ \eta_t\brac{\dF(w_t) - g_t}^\top\dF(w_t) + \frac{L\eta_t^2}{2}\norm{g_t}^2\; .
    \end{align*}
    Rearranging terms and dividing by $\eta_t$, we get
    \begin{align*}
        \norm{\dF(w_t)}^2 &\leq \frac{F(w_t) - F(w_{t+1})}{\eta_t} + \frac{L\eta_t}{2}\norm{g_t}^2 \\ &\qquad+ \brac{\dF(w_t) - g_t}^\top\dF(w_t)\; .
    \end{align*}
    Summing over $t=1,\ldots,T$, we have that
    \begin{align*}
        \sum_{t=1}^{T}{\norm{\dF(w_t)}^2} &\leq \sum_{t=1}^{T}{\frac{\Delta_t - \Delta_{t+1}}{\eta_t}} + \frac{L}{2}\sum_{t=1}^{T}{\eta_t\norm{g_t}^2} \\ &\quad+ \sum_{t=1}^{T}{\brac{\dF(w_t) - g_t}^\top\dF(w_t)} \; .
    \end{align*}
    Focusing on the first sum in the r.h.s., we have 
    \begin{align*}
        \sum_{t=1}^{T}{\frac{\Delta_t - \Delta_{t+1}}{\eta_t}} &\leq \frac{\Delta_1}{\eta_1} + \sum_{t=2}^{T}{\brac{\frac{1}{\eta_t} - \frac{1}{\eta_{t-1}}}\Delta_t} \leq \frac{\Delta_{\max}}{\eta_T},
    \end{align*}
    where the last inequality holds as $\eta_t$ is non-increasing. 
\end{proof}
To prove convergence to a stationary point, we bound each term in the r.h.s. of Lemma~\ref{lem:sgd_nonconvex_lemma}. The next lemma, which is commonly used in online learning, enables to bound the second term. 
\begin{lemma}[\citealp{auer2002adaptive}]\label{lem:lemma_gentile_auer}
    For any non-negative real numbers $a_1,\ldots,a_n$ it holds that
    \[
        \sum_{i=1}^{n}{\frac{a_i}{\sqrt{\sum_{j=1}^{i}{a_j}}}}\leq 2\sqrt{\sum_{i=1}^{n}{a_i}}\; .
    \]
\end{lemma}

\begin{theorem}\label{thm:nonconvex_convergence}
    Let $F:\reals^d\to\reals$ be an $L$-smooth function (Assumption~\ref{assump:smoothness}), bounded by $M$ (Assumption~\ref{assump:bounded_objective}), with bounded gradients (Assumption~\ref{assump:bounded_gradients}). Then, the iterates of Algorithm~\ref{alg:mlmc_adagrad} with $\alpha=1$ satisfy
    \[
        \bbE\sbrac{\frac{1}{T}\sum_{t=1}^{T}{\norm{\dF(w_t)}^2}}\leq \Ocaltilde\brac{\brac{M+L+G}G\sqrt{\frac{\taumix}{T}}}
    \]
\end{theorem}

\begin{proof}
    Plugging-in $\eta_t = (\sum_{k=1}^{t}{\norm{g_k}^2})^{-1/2}$ to the result from Lemma~\ref{lem:sgd_nonconvex_lemma}, we obtain
    \begin{align*}
        \sum_{t=1}^{T}{\norm{\dF(w_t)}^2} &\leq \Delta_{\max}\sqrt{\sum_{t=1}^{T}{\norm{g_t}^2}} \\ &\quad+ \frac{L}{2}\sum_{t=1}^{T}{\frac{\norm{g_t}^2}{\sqrt{\sum_{k=1}^{t}{\norm{g_k}^2}}}} \\ &\quad+ \sum_{t=1}^{T}{(\dF(w_t) - g_t)^\top\dF(w_t)}\; .
    \end{align*}
    Using Lemma~\ref{lem:lemma_gentile_auer} with $a_i=\norm{g_i}^2$ to bound the second term, we get that
    \begin{align}\label{eq:nonconvex_bias_adagrad_decomp}
        \sum_{t=1}^{T}{\norm{\dF(w_t)}^2} &\leq \brac{\Delta_{\max}+L}\sqrt{\sum_{t=1}^{T}{\norm{g_t}^2}} \nonumber \\ &\quad+ \sum_{t=1}^{T}{(\dF(w_t) - g_t)^\top\dF(w_t)}\; .
    \end{align}
    Under Assumption~\ref{assump:bounded_objective}, $\Delta_{\max}$ is bounded by $2M$. Taking expectation, the first term in the r.h.s. can be bounded using Jensen's inequality and the second order bound in Lemma~\ref{lem:mlmc_bounds}:
    \begin{align*}
        \bbE\sbrac{\brac{\Delta_{\max}+L}\sqrt{\sum_{t=1}^{T}{\norm{g_t}^2}}}&\leq \brac{2M+L}\sqrt{\sum_{t=1}^{T}{\bbE[\norm{g_t}^2]}} \\ &\leq\Ocaltilde\brac{\brac{M+L}G\sqrt{\taumix T}}. 
    \end{align*}
    The elements in the rightmost sum in Eq.~\eqref{eq:nonconvex_bias_adagrad_decomp} are bias terms, which resemble the bias terms in Theorem~\ref{thm:convex_convergence}; after taking expectation, they can be bounded similarly. Let $j_{\max}=\floor{\log{T}}$, and define $b_t(g) = (\dF(w_t) - g)^\top\dF(w_t)$. By the law of total expectation, Cauchy-Schwarz inequality, and Eq.~\eqref{eq:norm_bound}, we have that
    \begin{align*}
        \bbE\sbrac{b_t(g_t)} &= \bbE\sbrac{\bbE_{t-1}\sbrac{b_t(g_t)}} = \bbE[b_t(g_t^{j_{\max}})] \\ &\leq G\cdot\bbE\sbrac{\lVert\dF(w_t) - g_t^{j_{\max}}\rVert}\leq \Ocaltilde\brac{G^2\sqrt{\frac{\taumix}{T}}}.
    \end{align*}
    Therefore, the sum $\sum_{t=1}^{T}{\bbE[b_t(g_t)]}$ is bounded by $\Ocaltilde(G^2\sqrt{\taumix T})$. Overall, we get that $\bbE[\sum_{t=1}^{T}{\norm{\dF(w_t)}^2}]$ is bounded by $\Ocaltilde\brac{(M+L+G)G\sqrt{\taumix T}}$. Dividing by $T$ concludes the proof.
\end{proof}
\begin{corollary}\label{cor:nonconvex_convergence}
    Consider the setting in Theorem~\ref{thm:nonconvex_convergence}. Run Algorithm~\ref{alg:mlmc_adagrad} and let $\tilde{w}_T$ be an iterate chosen uniformly at random from $\{w_1,\ldots,w_T\}$. Then, 
    \vspace{-0.5em}
    \[
        \bbE\sbrac{\norm{\dF(\tilde{w}_T)}} \leq\Ocaltilde\brac{\sqrt{(M+L+G)G}\brac{\frac{\taumix}{T}}^{1/4}}\; .
    \]
    \vspace{-1.5em}
\end{corollary}
\begin{proof}
    By the definition of $\tilde{w}_T$, we have $\bbE[\norm{\dF(\tilde{w}_T)}^2]=\bbE[\frac{1}{T}\sum_{t=1}^{T}{\norm{\dF(w_t)}^2}]$. Using Jensen's inequality to bound $\bbE\sbrac{\norm{\dF(\tilde{w}_T)}}$ by $\sqrt{\bbE[\norm{\dF(\tilde{w}_T)}^2]}$ and injecting the result in Theorem~\ref{thm:nonconvex_convergence} finishes the proof.
\end{proof}

\section{Application to TD Learning with Linear Function Approximation}
In this section, we describe how MAG can be applied to TD learning with linear function approximation. We start with a concise background on TD learning in Section~\ref{sec:background_td}, and in Section~\ref{sec:TD_MAG} we provide a convergent TD algorithm whose convergence rate is $\Ocaltilde(\sqrt{\taumix/T})$ (see Theorem~\ref{thm:mag_td_convergence}); this rate has a better dependence on the mixing time compared to the $\Ocal(\taumix/\sqrt{T})$ rate of \citet{bhandari2018finite}. 
\vspace{-0.5em}
\subsection{Background on TD Learning}\label{sec:background_td}
In RL, an agent repeatedly interacts with an environment; the environment is usually modelled as a Markov Decision Process (MDP), which is defined by $(\Scal,\mathcal{A},\Pcal, \Rcal, \gamma)$, where $\Scal$ is the state space; $\mathcal{A}$ is the action space; $\Pcal:\mathcal{S}\times\mathcal{A}\times\mathcal{S}\to[0,1]$ is the transition kernel; $\Rcal:\mathcal{S}\times\mathcal{A}\to[-r_{\max}, r_{\max}]$ is the reward function; and $\gamma\in[0,1)$ is the discount factor. The agent employs a policy $\pi:\Scal\to\mathcal{A}$ that maps states to action, and by applying it in the MDP, we obtain samples from a Markov Reward Process (MRP) with induced transition and reward functions:
\begin{align*}
    &\Pcal^{\pi}(s,s')\coloneqq\Pcal(s, \pi(s), s'), \quad\forall s,s'\in\Scal, \\ &\mathcal{R}^{\pi}(s)\coloneqq\mathcal{R}(s, \pi(s)), \quad\forall s\in\Scal\; .
\end{align*}

We assume that the Markov chain underlying the MRP is ergodic with a unique stationary distribution $\mu$, and that the state space is finite and its size is $n=\abs{\Scal}$. 

Originally studied by~\citet{sutton1988learning}, TD learning is widely used in RL for estimating the value function $V^{\pi}(s)\coloneqq\bbE_{\distributed{s_t}{\Pcal^{\pi}}}[\sum_{t=0}^{\infty}{\gamma^t\Rcal^{\pi}(s_t)}|s_0=s]$ by enforcing the Bellman equation:
\[
    V^{\pi}(s) = \Rcal^{\pi}(s) + \gamma\sum_{s'\in\Scal}{\Pcal^{\pi}(s'|s)V^{\pi}(s')},\quad\forall s\in\Scal\; .
\]
Since in modern problems the number of states $n$ is extremely large, it is common to use some parametric function to approximate the value function. Here, we focus on linear function approximation~\citep{tsitsiklis1997analysis}: We consider some bounded feature mapping $\phi:\Scal\to\reals^d$ such that $\norm{\phi(s)}\leq 1, \forall s\in\Scal$ and aim to find a vector $\theta\in\reals^d$ for which the value function approximately satisfies $V^\pi(s)\approx V_{\theta}(s)\coloneqq\phi(s)^\top\theta$. To put it more formally, let $M = \text{diag}(\mu(s_1),\ldots,\mu(s_n))$ denote the diagonal matrix whose entries correspond to the stationary distribution $\mu$; our goal is to find $\theta$ that minimizes the weighted mean squared error, defined by 
\begin{equation}\label{eq:td_objective}
    \norm{V^{\pi} - V_{\theta}}_{M}^2 \coloneqq \sum_{s\in\Scal}{\mu(s)\brac{V^{\pi}(s) - V_{\theta}(s)}^2}\; .
\end{equation}
Denoting a sample from the MRP at time $t$ by $z_t = \brac{s_t,r_t=\calig{R}^{\pi}(s_t, s_t'), s_t'}$, the update rule of TD learning with linear function approximation is given by
\begin{equation}\label{eq:td_linear_update}
    \theta_{t+1} \coloneqq \theta_t + \eta_t g(\theta_t; z_t)\; ,
\end{equation}
where
\begin{equation}\label{eq:td_gradient}
    g(\theta; z_t) = \brac{r_t + \gamma\phi(s_t')^\top\theta - \phi(s_t)^\top\theta}\phi(s_t)\; ,
\end{equation}
and $\eta_t>0$ is the learning rate. The vector $g(\theta;z_t)$ is sometimes referred to as a \textit{semi-gradient}. Let $\bar{g}(\theta)$ denote the expectation of $g(\theta;z_t)$ when $s_t\sim\mu$ and $s_t'\sim\calig{P}^\pi(\cdot|s_t)$, which characterizes the steady-state behavior of $g(\theta;z_t)$; that is,
\begin{equation*}
    \bar{g}(\theta)\coloneqq\underset{\begin{subarray}{c}\distributed{s}{\mu} \\\distributed{s'}{\Pcal^{\pi}(\cdot|s)}\end{subarray}}{\bbE}\sbrac{\brac{\Rcal^{\pi}(s,s') + \gamma\phi(s')^\top\theta - \phi(s)^\top\theta}\phi(s)}.
\end{equation*}
The theory of stochastic approximation guarantees that, under appropriate selection of learning rate, the iterates defined by Eq.~\eqref{eq:td_linear_update} converge to a limit point $\theta^*$ that satisfies $\bar{g}(\theta^*)=0$; this point enjoys a favorable theoretical guarantee in terms of minimizing the objective in Eq.~\eqref{eq:td_objective} (see Theorem 1 of \citealp{tsitsiklis1997analysis}). Unfortunately, stochastic approximation only ensures asymptotic convergence and do not provide any finite-time guarantees. Recently, \citet{bhandari2018finite} drew a connection between TD learning and SGD, which enabled them to give convergence bounds by mirroring standard SGD arguments.
A key lemma in their analysis establishes a property that mirrors the gradient inequality from convex optimization.
\begin{lemma}[\citealt{bhandari2018finite}, Lemma 3]\label{lem:bhandari} For any $\theta\in\reals^d$,
\[
    \bar{g}(\theta)^\top(\theta^*-\theta)\geq (1-\gamma)\norm{V_{\theta^*} - V_{\theta}}^2_M\; .
\]
\end{lemma}
\subsection{TD Learning with MAG}\label{sec:TD_MAG}
In this section we describe how to adapt MAG to TD learning with linear function approximation and state a convergence bound for the resulting algorithm. 

Similarly to Section~\ref{sec:MAG}, we use $s_t^{(i)}$ to denote the $i$-th state in the $t$-th iteration. Accordingly, we denote the corresponding~\textit{state}-\textit{reward}-\textit{next\hspace{0.2em}state} triplet by $z_t^{(i)}=(s_t^{(i)}, r_t^{(i)}, s_t'^{(i)})$, where $r_t^{(i)} = \calig{R}^{\pi}(s_t^{(i)}, s_t'^{(i)})$ and $s_t'^{(i)}\sim\mathcal{P}^{\pi}(\cdot|s_t^{(i)})$. Considering the semi-gradient in Eq.~\eqref{eq:td_gradient} and slightly abusing notation, we 
define $g_t^{j}~=~\frac{1}{2^{j}}\sum_{i=1}^{2^j}{g(\theta_t;z_t^{(i)})}$ and use the MLMC estimator defined in Eq.~\eqref{eq:mlmc_grad_estimator} for estimating  $\bar{g}(\theta_t)$. 

Following \citet{bhandari2018finite}, we consider the projected TD update rule: letting $\K=\cbrac{\theta\in\reals^d: \norm{\theta}\leq R}$ for some $R>0$ such that $\norm{\theta^*}\leq R$, our TD algorithm is defined by the following update rule:
\begin{equation}\label{eq:mag_td}
    \theta_{t+1} = \proj{\K}{\theta_t + \eta_t g_t}; \quad\eta_t = \frac{\sqrt{2}R}{\sqrt{\sum_{k=1}^{t}{\norm{g_k}^2}}}\; ,
\end{equation}
where $g_t$ is the MLMC estimator described above. 

Note that proving the convergence of MAG in the convex setting requires Assumptions~\ref{assump:bounded_gradients} and \ref{assump:bounded_domain} to hold, i.e., bounded (semi-)gradients and bounded domain. Here, the domain is bounded by $D=2R$ due to the projection step, and since the features and rewards are bounded, it holds that $\lVert g(\theta_t; z_t^{(i)})\rVert\leq r_{\max} + 2R,\enskip\forall i\in\naturals, t\in\sbrac{T}$.

\begin{theorem}\label{thm:mag_td_convergence}
    Let $G=r_{\max} + 2R$. The iterates defined by Eq.~\eqref{eq:mag_td} satisfy
    \[
        \mathbb{E}\sbrac{\norm{V_{\theta^*} - V_{\bar{\theta}_T}}_{M}^2} \leq \tilde{\mathcal{O}}\brac{\frac{GR\sqrt{\taumix}}{(1-\gamma)\sqrt{T}}}\; ,
    \]
    where $\bar{\theta}_{T}\coloneqq\frac{1}{T}\sum_{t=1}^{T}{\theta_t}$. 
\end{theorem}
The proof of this theorem closely follows the proof of Theorem~\ref{thm:convex_convergence} with key difference that we use Lemma~\ref{lem:bhandari} instead of the gradient inequality. We defer the proof to Appendix~\ref{app:proof_thm_mag_td}. Theorem~\ref{thm:mag_td_convergence} implies that our method enjoys a convergence rate that is $\sqrt{\taumix}$ times faster than the $\Ocal\brac{\frac{G^2\taumix}{(1-\gamma)\sqrt{T}}}$ rate of \citet{bhandari2018finite}. In Appendix~\ref{app:linear_mix_time_example} we give a motivating example, in which we show that the mixing time of a Markov chain can grow linearly with the number of states; in such problems, this $\sqrt{\taumix}$ factor improvement is substantial.

\section{Experiments}\label{sec:experiments}
We demonstrate the performance of our approach on a synthetic linear regression problem with Markovian data. In all our experiments, we report the average performance over $5$ random seeds as well as the $95\%$ confidence interval. 
\begin{figure}[t]
    \begin{center}
        \includegraphics[width=\columnwidth]{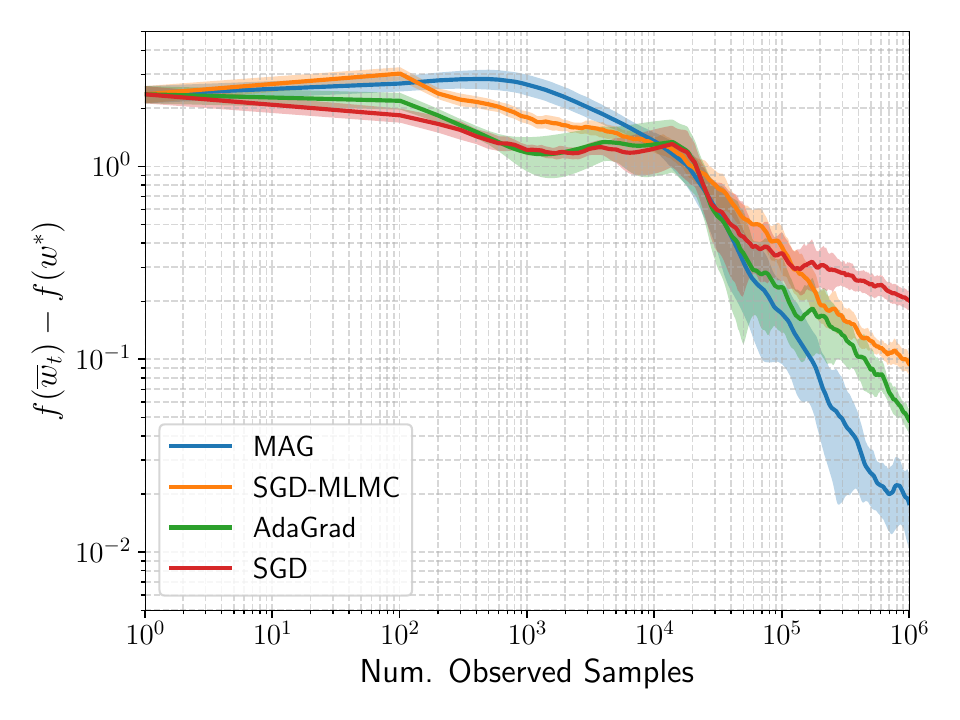}
        \caption{Suboptimality of the average iterate $\bar{w}_t = \frac{1}{t}\sum_{k=1}^{t}{w_k}$ as a function of the number of observed Markovian samples, when the transition parameter is $p=10^{-4}$ ($\taumix\approx 10^{4}$).}
        \label{fig:performance_comparison}
    \end{center}
\end{figure}

\begin{figure}[t]
    \begin{center}
        \includegraphics[width=\columnwidth]{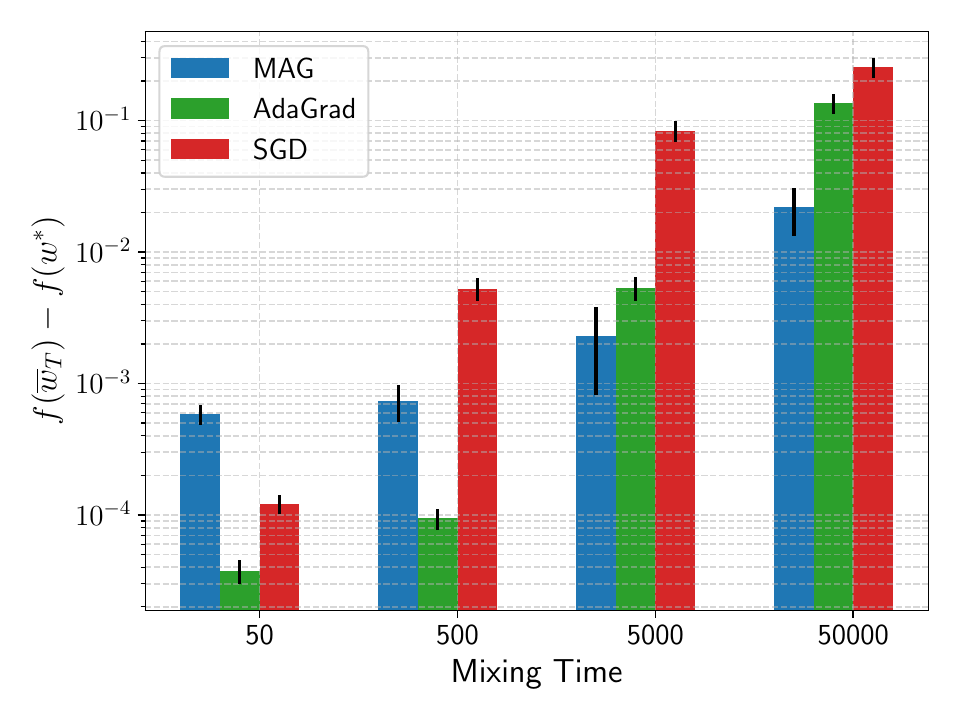}
        \caption{Suboptimality of the last average iterate $\bar{w}_T$ after observing $T=5\cdot 10^{6}$ samples for different values of mixing time.}
        \label{fig:mixing_time_ablation}
    \end{center}
\end{figure}
We consider a symmetric $2$-state Markov chain, for which the probability of transitioning between states is $p\in(0,1)$. 
For each state $i\in\cbrac{1,2}$, we randomly choose $w_i^{\dag}\in\reals^d$ and draw $X_{i}\in\reals^{n\times{d}}$ according to $\mathcal{N}(0,I)$. Then, we set $y_{i} = X_{i}w_i^{\dag} + \epsilon$, where $\epsilon$ is a random noise vector~$\sim\mathcal{N}(0, 10^{-3})$. Since the chain is symmetric, its stationary distribution is uniform, $\mu=\brac{\frac{1}{2}, \frac{1}{2}}$, and therefore our objective given by
\[
    \min_{\norm{w}\leq r}{\frac{1}{4n}\norm{
    \left(\begin{array}{c}
        X_1 \\
        X_2
    \end{array}\right)w - \left(\begin{array}{c}
        y_1 \\
        y_2
    \end{array}\right)
    }^2}\; . 
\]
However, in each step we only observe data associated with the current state of the chain. This problem is closely related to the decentralized optimization framework of \citet{ram2009incremental, johansson2010randomized}. 

We compared our algorithm with: \textbf{(i)} \textbf{SGD} with decreasing learning rate $\eta_t = 1/\sqrt{t}$; \textbf{(ii)} \textbf{AdaGrad} with standard gradient estimator; and \textbf{(iii)} \textbf{SGD-MLMC}, which uses the MLMC estimator and a non-adaptive learning rate, $\eta_t=1/\sqrt{t}$. For both MAG and AdaGrad we used $\eta_t = (\sum_{k=1}^{t}{\lVert g_k\rVert^2})^{-1/2}$. Since there exists some learning rate (that depends on the mixing time) for which SGD guarantees optimal performance~\cite{duchi2012ergodic}, we did not tune the scaling parameter of any learning rate. 

In practice, we observed that the multiplicative $2^{J}$ factor in the MLMC estimator resulted in high magnitude gradients which in turn significantly reduced the learning rate of AdaGrad and slowed down future progress. Therefore, instead of drawing $\distributed{J}{\text{Geom}(\frac{1}{2})}$, we used a truncated geometric distribution such that $\prob(J=j)\propto 2^{-j}, j\in\sbrac{K}$ for some fixed $K$. We used $K=5$, which was not optimally tuned. 

In Figure~\ref{fig:performance_comparison} we present the performance of the methods for $n=250, d=100$ and $p=10^{-4}$.  We observe that MAG outperforms other methods. Specifically, when we compare MAG with AdaGrad and SGD-MLMC with SGD, where in both cases the learning rate is set similarly, the method that employs the MLMC estimator performs better; this comparison validates the effectiveness of our proposed MLMC estimator for problems with Markovian data.

To study the effect of the mixing time on performance, we evaluated the methods for different values of $\taumix$.
We control the mixing time by setting the transition parameter $p$; when $p$ is small, it can be easily shown that $\taumix=\Theta(1/p)$ (see Appendix~\ref{app:proof_thm_convex_case}). In Figure~\ref{fig:mixing_time_ablation} we present the final performance of the methods after observing $5\cdot 10^{6}$ samples for different magnitudes of mixing time (we omit the results for SGD-MLMC since for all the tested values of $\taumix$, MAG obtained better performance). We observe that MAG outperforms other methods when $\taumix$ is large. One possible explanation for its inferiority in the small mixing time regime is that the benefit of low bias gradients is overshadowed by a high variance. We also note that the performance of MAG degrades more gracefully when $\taumix$ increases.

\section{Conclusion}
We study stochastic optimization problems with data arriving from a Markov chain. Although performance of algorithms in this setting depends on the mixing time of the chain, we propose a first-order method that does not rely on prior knowledge of the mixing time, yet still obtains optimal convergence rate when applied to convex problems. This method relies on combining an MLMC gradient estimator with an adaptive learning rate. We show that our adaptive algorithm can also be used for finding stationary points in non-convex problems and for estimating the value function in the context of TD learning. Finally, we validate our theoretical findings on a linear regression task with Markovian data, demonstrating the effectiveness of our method.

\section*{Acknowledgement}
The authors would like to thank Rotem Aviv and Guy Raveh for helpful discussions and for comments on the paper draft, and the reviewers for their useful comments. This research was partially supported by Israel PBC-VATAT, by the Technion Center for Machine Learning and Intelligent Systems (MLIS) and by the Israel Science Foundation (grant No. 447/20).

\bibliography{icml_refs}
\bibliographystyle{icml2022}





\appendix
\onecolumn

\section{Gradient Estimation with Markovian Samples}
In this section, we derive a concentration bound for estimating the gradient of some function using the average of stochastic gradients based on samples from a Markov chain. Consider the Markov chain $z_1,z_2,\ldots$ and some $w\in\K$. We construct the following estimator for the gradient of $F$ at $w$ using $N$ samples from the chain:
\begin{equation}\label{eq:avg_gradinet_est}
    g^{N}(w)\coloneqq\frac{1}{N}\sum_{i=1}^{N}{\nabla f(w; z_{i})}\; .
\end{equation}
First, we assume that the chain is stationary and that $w$ is independent of the chain. Under these assumptions, we derive a concentration bound for $\norm{g^N(w) - \dF(w)}$. Using this result, we will later obtain a high probability bound for the case where the chain is not necessarily stationary, and $w$ might depend on it.

We start with the following result which bounds the probability of an equally spaced subsample of the chain being different than its coupled i.i.d. counterpart.
\begin{lemma}[\citealp{bresler2020least}, Lemma 3]\label{lem:coupling_bresler}
    Let $z_1,z_2,\ldots$ be a stationary finite state Markov chain with stationary distribution $\mu$, and let $K,n\in\mathbb{N}$. Then, we can couple $(z_1,z_{K+1},\ldots,z_{(n-1)K+1})$ and $(\tilde{z}_1,\tilde{z}_{K+1},\ldots,\tilde{z}_{(n-1) K+1})\sim\mu^{\otimes{n}}$ such that:
    \[
        \prob\brac{(z_1,z_{K+1},\ldots,z_{(n-1)K+1})\neq(\tilde{z}_1,\tilde{z}_{K+1},\ldots,\tilde{z}_{(n-1) K+1})}\leq (n-1)\dmix{K}\; ,
    \]
    where $\dmix{\cdot}$ is defined in Eq.~\eqref{eq:d_mix}. 
\end{lemma}
Another useful result we build on is a concentration bound for the average of i.i.d vectors.
\begin{lemma}[\citealp{kakade2010concentration}, Theorem 1.1]\label{lem:kakade_concentration_vec}
    Let $x\in\reals^d$ be a random vector. Assume that $x_1,\ldots,x_n\in\reals^d$ are sampled i.i.d. from some distribution such that $\bbE x_i = x,\forall i\in\sbrac{n}$, and $\norm{x_i}\leq M$ almost surely for some $M>0$. Then, with probability at least $1-\delta$,
    \[
        \norm{\frac{1}{n}\sum_{i=1}^{n}{x_i} - x}\leq \frac{6M}{\sqrt{n}}\brac{1+\sqrt{\log{(1/\delta)}}}\; .
    \]
\end{lemma}
Equipped with these two lemmas, we are now able to derive a concentration bound for $g^N(w)$.
\begin{lemma}[Concentration Inequality for Stationary Markov Chain Samples]\label{lem:concentration_stationary_markov}
    Let $z_1,z_2,\ldots$ be a stationary finite state Markov chain, and consider $g^{N}(w)$ defined in Eq.~\eqref{eq:avg_gradinet_est}. Suppose $\norm{\nabla f(w;z)}\leq G$ for every $w\in\K, z\in\Omega$. Let $K\in\sbrac{N}$ and set $n=\floor{N/K}$. Then, for every $\delta>K(n-1)\dmix{K}$, with probability at least $1-\delta$, the following holds for every fixed $w\in\K$:
    \[
        \norm{g^N(w) - \nabla F(w)} \leq\frac{6G}{\sqrt{n}}\brac{1 + \sqrt{\log{(K/\delta')}}} +  \frac{2GK}{N}\; ,
    \]
    where $\delta'=\delta-K(n-1)\dmix{K}$.
\end{lemma}
\begin{proof}
    We divide $z_1,\ldots,z_N$ into two sets: $S_1 = (z_1,\ldots,z_{n K})$ and $S_2=(z_{nK+1},\ldots,z_N)$. We then divide $S_1$ into $K$ blocks of length $n$ each, such that each block contains samples that are approximately independent (for large enough $K$) and use concentration arguments combined with Lemma~\ref{lem:coupling_bresler}. For the remainder $S_2$, we use a trivial coarse bound. 
    
    Let $w\in\K$, and for every $k\in\sbrac{K}$ let us define the following subsample: $\Z_k = \brac{z_{k}, z_{K+k}, \ldots, z_{(n-1)K+k}}$. Similarly, we define the coupled sequences $\tilde{\Z}_k =(\tilde{z}_k,\tilde{z}_{K+k},\ldots,\tilde{z}_{(n-1)K+k})\sim\mu^{\otimes{n}}$ for every $k\in\sbrac{K}$. Also define the following set of events for $k\in\sbrac{K}$: $\mathcal{\E}_k = \cbrac{\Z_k =\tilde{\Z}_k}$, and denote $d(\Z_k) = \norm{\frac{1}{n}\sum_{i=1}^{n}{\nabla f(w;z_{(i-1)K+k})} - \nabla F(w)}$. Let $\epsilon>0$. For every $k\in\sbrac{K}$, we have:
    \begin{align*}
        \prob\brac{d(\Z_k)>\epsilon} &= \prob\brac{\cbrac{d(\Z_k)>\epsilon}\cap \E_k} + \prob\brac{\cbrac{d(\Z_i)>\epsilon}\cap\E_k^c} \\ &\leq \prob(d(\tilde{\Z}_k)>\epsilon) + \prob\brac{\E_k^c} \\ &\leq \exp\brac{-\brac{\frac{\epsilon\sqrt{n}}{6G}-1}^2} + (n-1)\dmix{K}\; ,
    \end{align*}
    where in the last inequality we used both Lemma~\ref{lem:kakade_concentration_vec}, since the elements in $\tilde{\Z}_k$ are i.i.d and $\norm{\df(w;z)}\leq G$ almost surely, and Lemma~\ref{lem:coupling_bresler} to bound the probability of $\E_k^c$. Using this result, we can bound the probability of  $\norm{\frac{1}{nK}\sum_{i=1}^{nK}{\df(w;z_{i})}- \dF(w)}$ being greater than $\epsilon$ as follows:
    \begin{align*}
        \prob\brac{\norm{\frac{1}{nK}\sum_{i=1}^{nK}{\df(w;z_i)}- \dF(w)}>\epsilon} &= \prob\brac{\norm{\frac{1}{K}\sum_{k=1}^{K}{\brac{\frac{1}{n}\sum_{i=1}^{n}{\df(w;z_{(i-1)K+k})} - \dF(w)}}}>\epsilon} \\ &\overset{\dag}{\leq} \prob\brac{\frac{1}{K}\sum_{k=1}^{K}{\norm{\frac{1}{n}\sum_{i=1}^{n}{\df(w;z_{(i-1)K+k})} - \dF(w)}}>\epsilon} \\ &= \prob\brac{\frac{1}{K}\sum_{k=1}^{K}{d(\Z_k)}>\epsilon} \\ &\overset{\ddag}{\leq} \sum_{k=1}^{K}{\prob\brac{d(\Z_k)>\epsilon}} \\ &\leq K\exp\brac{-\brac{\frac{\epsilon\sqrt{n}}{6G}-1}^2} + K(n-1)\dmix{K}\; ,
    \end{align*}
    where $\dag$ holds by the triangle inequality, and $\ddag$ holds by the union bound and the fact that one of $d(\Z_k)$ must surpass $\epsilon$ for their average to surpass $\epsilon$. Setting the r.h.s. to be $\delta$, we have with probability at least $1-\delta$ that
    \begin{equation}\label{eq:hp_bound_delta_'}
        \norm{\frac{1}{nK}\sum_{i=1}^{nK}{\df(w;z_i)}- \dF(w)} \leq \frac{6G}{\sqrt{n}}\brac{1 + \sqrt{\log{(K/\delta')}}}\; ,
    \end{equation}
    where $\delta'=\delta-K(n-1)\dmix{K}$. Finally, let us write 
    \begin{align*}
        \norm{g^N(w) - \nabla F(w)} &= \norm{\frac{1}{N}\sum_{i=1}^{N}{\df(w;z_i)} - \dF(w)} \\ &= \norm{\frac{1}{nK}\sum_{i=1}^{nK}{\df(w;z_i)} - \dF(w) + \brac{\frac{1}{N} - \frac{1}{nK}}\sum_{i=1}^{nK}{\df(w;z_i)} + \frac{1}{N}\sum_{i=nK+1}^{N}{\nabla f(w;z_i)}} \\ &\leq \norm{\frac{1}{nK}\sum_{i=1}^{nK}{\df(w;z_i)}- \dF(w)} + \brac{\frac{1}{nK} - \frac{1}{N}}{\sum_{i=1}^{nK}{\norm{\df(w;z_i)}}} + \frac{1}{N}\sum_{i=nK+1}^{N}{\norm{\df(w;z_i)}} \\ &\leq \norm{\frac{1}{nK}\sum_{i=1}^{nK}{\df(w;z_i)}- \dF(w)} + \frac{2G(N-nK)}{N} \; ,
    \end{align*}
    where in the last inequality we used the boundness of each $\norm{\df(w;z)}$. Since $nK = \floor{N/K}K\geq (N/K-1)K = N-K$, we have that $N-nK\leq K$, and the last term in the r.h.s. is bounded $2GK/N$. Combining this with the high probability bound in Eq.~\eqref{eq:hp_bound_delta_'} concludes the proof. 
\end{proof}

The above lemma establishes a concentration bound for the case where the chain $z_1,z_2,\ldots$ is stationary and independent of $w$. Since we are interested in analyzing the iterates of algorithms that use gradient averaging based on Markovian samples, from now on we consider $w_t$ that depends on previous samples from the chain. Moreover, we no longer assume that the Markov chain $z_{t}^{(1)}, z_{t}^{(2)},\ldots$ is stationary. In the following lemma we take $w_t$ that is measurable with respect to $\F_{t-1}$, and consider the average of $M$ stochastic gradients based on Markovian samples that are delayed by $K$ steps. That is, we are interested in $\frac{1}{M}\sum_{i=1}^{M}{\df(w_t;z_{t}^{(K+i)})}$ and provide a high probability bound for its distance from the true gradient.
\begin{lemma}\label{lem:concentraion_at_time_t}
    Suppose $\norm{\nabla f(w;z)}\leq G$ for every $w\in\K, z\in\Omega$. Let $M\in\naturals$, $K\in\sbrac{M}$, and $n=\floor{M/K}$. Then, for every $\delta>(K(n-1)+1)\dmix{K}$, the following holds for every $w_t\in\K$ measurable w.r.t. $\F_{t-1}$: 
    \[
        \prob_{t-1}\brac{\norm{\frac{1}{M}\sum_{i=1}^{M}{\df(w_t; z_t^{(K+i)}) - \dF(w_t)}}\leq \frac{6G}{\sqrt{n}}\brac{1 + \sqrt{\log{(K/\delta')}}} +  \frac{2GK}{M}}\geq 1-\delta\; \; ,
    \]
    where $\delta' = \delta - (K(n-1)+1)\dmix{K}$.
\end{lemma}
\begin{proof}
    Consider a time-homogenous Markov chain $\hat{z}_{t}^{(K)}, \hat{z}_{t}^{(K+1)},\ldots,\hat{z}_{t}^{(K+M)}$ with $\hat{z}_{t}^{(K)}\sim\mu$ and with the same transition matrix $P$ as the original chain. Moreover, assume that $\hat{z}_t^{(K)}$ is independent of the original chain.
    The fact that the two chains have the same transition matrix implies the following,
    \begin{equation}\label{eq:star}
        \prob\brac{z_t^{(K+1)}, \ldots, z_{t}^{(K+M)}\mid z_t^{(K)}=z} = \prob(\hat{z}_t^{(K+1)}, \ldots, \hat{z}_{t}^{(K+M)}\mid\hat{z}_t^{(K)}=z), \quad \forall z\in\Omega\; . 
    \end{equation}
    Let $\epsilon>0$ and let us denote the following two events,
    \[
        \A = \cbrac{\norm{\frac{1}{M}\sum_{i=1}^{M}{\df(w_t; z_t^{(K+i)}) - \dF(w_t)}}\geq\epsilon}\enskip , \quad\hat{\A} = \cbrac{\norm{\frac{1}{M}\sum_{i=1}^{M}{\df(w_t; \hat{z}_t^{(K+i)}) - \dF(w_t)}}\geq\epsilon}\; .
    \]
    Since $w_t$ is measurable w.r.t. $\F_{t-1}$, given $\F_{t-1}$, the only stochasticity in $\A$ is due to $z_{t}^{(K+1)},\ldots,z_t^{(K+M)}$. Moreover, by the Markov property, given $z_{t}^{(K)}$, the distribution of $z_{t}^{(K+1)},\ldots,z_t^{(K+M)}$ is independent of $\F_{t-1}$. Combining these two observations, we conclude that $\prob_{t-1}(\A|z_t^{(K)}=z)$ only depends on the conditional probability $\prob\brac{z_t^{(K+1)}, \ldots, z_{t}^{(K+M)}\mid z_t^{(K)}=z}$. Therefore, according to Eq.~\eqref{eq:star}, we have $\prob_{t-1}(\A|z_t^{(K)}=z) = \prob_{t-1}(\hat{\A}|\hat{z}_t^{(K)}=z)$. 
    By the law of total probability we have
    \begin{align*}
        \prob_{t-1}(\A) &= \sum_{z\in\Omega}{\prob_{t-1}(\A|z_t^{(K)}=z)\cdot\prob_{t-1}(z_t^{(K)}=z)} \\ &= \sum_{z\in\Omega}{\prob_{t-1}(\hat{\A}|\hat{z}_t^{(K)}=z)\cdot \prob_{t-1}(z_t^{(K)}=z)} \\ &= \sum_{z\in\Omega}{\prob_{t-1}(\hat{\A}|\hat{z}_t^{(K)}=z)\cdot\prob(\hat{z}_t^{(K)}=z)} +  \sum_{z\in\Omega}{\prob_{t-1}(\hat{\A}|\hat{z}_t^{(K)}=z)\brac{\prob_{t-1}(z_t^{(K)}=z) - \prob(\hat{z}_t^{(K)}=z)}} \\ &= \sum_{z\in\Omega}{\prob_{t-1}(\hat{\A}|\hat{z}_t^{(K)}=z)\cdot\prob_{t-1}(\hat{z}_t^{(K)}=z)} +  \sum_{z\in\Omega}{\prob_{t-1}(\hat{\A}|\hat{z}_t^{(K)}=z)\brac{\prob_{t-1}(z_t^{(K)}=z)- \mu(z)}} \\ &\overset{\dag}{\leq} \prob_{t-1}(\hat{\A}) + \sum_{z\in\Omega}{\abs{\prob_{t-1}(\hat{\A}|\hat{z}_t^{(K)}=z)} \big\lvert \prob_{t-1}(z_t^{(K)}=z) - \mu(z)\big\rvert} \\ &\leq \prob_{t-1}(\hat{\A}) + \sum_{z\in\Omega}{\big\lvert \prob_{t-1}(z_t^{(K)}=z) - \mu(z)\big\rvert} \\ &= \prob_{t-1}(\hat{\A}) + \tv{\prob_{t-1}(z_t^{(K)})}{\mu} \\ &\leq \prob_{t-1}(\hat{\A}) + \dmix{K}\; ,
    \end{align*}
    where in $\dag$ we used the fact that $\distributed{\hat{z}_t^{(K)}}{\mu}$ and the law of total probability, and the last inequality holds by the definition of $\textup{d}_\textup{mix}$ in Eq.~\eqref{eq:d_mix}. Note that $w_t$ is measurable w.r.t. $\F_{t-1}$; $\hat{z}_t^{(K+1)},\ldots,\hat{z}_t^{(K+M)}$ is a stationary Markov chain; and $K\leq M$. Thus, we can use Lemma~\ref{lem:concentration_stationary_markov} to bound $\prob_{t-1}(\hat{\A})$ as follows,
    \[
        \prob_{t-1}(\hat{\A}) \leq K\exp\brac{-\brac{\frac{\sqrt{n}}{6G}\brac{\epsilon-\frac{2GK}{M}} - 1}^2} + K(n-1)\dmix{K}\; ,
    \]
    which implies that
    \[
        \prob_{t-1}(\A) \leq K\exp\brac{-\brac{\frac{\sqrt{n}}{6G}\brac{\epsilon-\frac{2GK}{M}} - 1}^2} + (K(n-1)+1)\dmix{K}\; .
    \]
    Setting the r.h.s. to be $\delta$ and writing $\epsilon$ in terms of $\delta$ concludes the proof.
\end{proof}

Using this result, we can finally prove a high probability bound for the average of stochastic gradients without delaying the Markovian samples. The technique for proving this bound combines Lemma~\ref{lem:concentraion_at_time_t} with a crude bound for the stochastic gradients computed using the first $K$ samples.

\begin{lemma}[Concentration inequality for Markov Chain Samples]\label{lem:concentraion_mc}
Let $K, N\in\naturals$ such that $N\geq 2K$. Assume that $\norm{\df(w;z)}\leq G$ for every $w\in\K, z\in\Omega$. Then, for every $\delta>N\dmix{K}$, the following holds for every $w_t\in\K$ measurable w.r.t. $\F_{t-1}$:
    \begin{equation*}
        \prob_{t-1}\brac{\norm{\frac{1}{N}\sum_{i=1}^{N}{\df(w_t;z_t^{(i)})} - \dF(w_t)}\leq 12G\sqrt{\frac{K}{N}}\brac{1 + \sqrt{\log{(K/\tilde{\delta})}}} + \frac{6GK}{N}}\geq 1-\delta\; ,
    \end{equation*}
    where $\tilde{\delta} = \delta-N\dmix{K}$.
\end{lemma}
\begin{proof}
    Let us write,
    \begin{align*}
        \norm{\frac{1}{N}\sum_{i=1}^{N}{\df(w_t;z_t^{(i)})} - \nabla F(w_t)} &\leq \frac{1}{N}\norm{\sum_{i=1}^{K}{(\df(w_t;z_t^{(i)}) - \dF(w_t))}} + \norm{\frac{1}{N}\sum_{i=K+1}^{N}{(\df(w_t;z_t^{(i)}) - \dF(w_t))}} \\ &\leq \frac{2GK}{N} + \frac{N-K}{N}\norm{\frac{1}{N-K}\sum_{i=1}^{N-K}{\df(w_t;z_t^{(K+i)}) - \dF(w_t)}}\\ &\leq \frac{2GK}{N} + \norm{\frac{1}{N-K}\sum_{i=1}^{N-K}{\df(w_t;z_t^{(K+i)}) - \dF(w_t)}} \; .
    \end{align*}
    Using Lemma~\ref{lem:concentraion_at_time_t} with $M=N-K\geq K$, we can bound the second term: Given $\F_{t-1}$, for every $\delta>(K(\floor{\frac{N-K}{K}}-1)+1)\dmix{K}$, with probability at least $1-\delta$ we have
    \[
        \norm{\frac{1}{N-K}\sum_{i=1}^{N-K}{\df(w_t;z_t^{(K+i)}) - \dF(w_t)}} \leq \frac{6G}{\sqrt{\left\lfloor\frac{N-K}{K}\right\rfloor}}\brac{1+\sqrt{\log{(K/\delta')}}} + \frac{2GK}{N-K}\; ,
    \]
    where $\delta'=\delta-(K(\floor{\frac{N-K}{K}}-1)+1)\dmix{K}$. Since $N\geq 2K$ we have $N-K\geq N/2$, and since for every $x\geq 1$ it holds that $\floor{x}\geq x/2$, we have $\floor{\frac{N-K}{K}}\geq \floor{\frac{N/2}{K}}\geq \frac{N}{4K}$, which results in
    \[
         \norm{\frac{1}{N-K}\sum_{i=1}^{N-K}{\df(w_t;z_t^{(K+i)}) - \dF(w_t)}} \leq 12G\sqrt{\frac{K}{N}}\brac{1 + \sqrt{\log{(K/\delta')}}} + \frac{4GK}{N}\; .
    \]
    Moreover, note that 
    \[
        \brac{K\brac{\left\lfloor\frac{N-K}{K}\right\rfloor-1}+1}\dmix{K}\leq \brac{K\brac{\frac{N-K}{K}-1}+1}\dmix{K} = (N-2K+1)\dmix{K}\leq N\dmix{K}\; ,
    \]
    which implies that $\delta'\geq\delta-N\dmix{a}$. Combining the above results, we get that
    \[
        \prob_{t-1}\brac{\norm{\frac{1}{N}\sum_{i=1}^{N}{\df(w_t;z_t^{(i)})} - \nabla F(w_t)}\leq 12G\sqrt{\frac{K}{N}}\brac{1 + \sqrt{\log{(K/\tilde{\delta})}}} + \frac{6GK}{N}}\geq 1-\delta\; ,
    \]
    where $\tilde{\delta} = \delta-N\dmix{a}$.
\end{proof}

Following the high probability bound above, we can now derive a bound on the expectation and second moment of $\norm{\frac{1}{N}\sum_{i=1}^{N}{\df(w_t;z_t^{(i)})-\dF(w_t)}}$. Intuitively, we take $K$ to be proportional to $\taumix\log{T}$ for some $T\geq N$, which enables to choose a small confidence parameter $\delta$ and still provides satisfactory bounds. 

\begin{lemma}\label{lem:expectation_and_second_order_bound}
    Assume that $\norm{\df(w;z)}\leq G$ for every $w\in\K, z\in\Omega$ and let $g_t^{N}\coloneqq\frac{1}{N}\sum_{i=1}^{N}{\df(w_t;z_t^{(i)})}$. Moreover, for some integer $T\in\naturals$, let $K=\taumix\ceil{2\log{T}}$. Then, for every $N\in\sbrac{T}$ and every $w_t$ measurable w.r.t. $\F_{t-1}$, the following holds:  
    \[
        \bbE\sbrac{\norm{g_t^N - \nabla F(w_t)}} \leq \Ocaltilde\brac{G\sqrt{\frac{K}{N}}}\; ,
    \]
    and
    \[
        \bbE\sbrac{\norm{g_t^{N} - \dF(w_t)}^2}\leq \Ocaltilde\brac{\frac{G^2 K}{N}}\; .
    \]
\end{lemma}
\begin{proof}
    We consider two different cases for $N$. \\
    
    \noindent\underline{Case 1: $N<2K$.} We can trivially bound $\norm{g_t^{N} - \dF(w_t)}$ by $2G$ using the triangle inequality. Since $N<2K$, it holds that $1\leq \sqrt{2K/N}$, and therefore $\bbE\sbrac{\norm{g_t^N - \nabla F(w_t)}}\leq 2G\leq\Ocaltilde(G\sqrt{K/N})$. Since $\norm{g_t^{N} - \dF(w_t)}^2$ is trivially bounded by $4G^2$, we similarly obtain the bound for the second moment. \\
    
    \noindent\underline{Case 2: $N\geq 2K$.} In this case,
    we bound the conditional expectations $\bbE_{t-1}\sbrac{\norm{g_t^{N} - \dF(w_t)}}$ and $\bbE_{t-1}[\lVert g_t^{N} - \dF(w_t)\rVert^2]$, which implies a bound on the expectations themselves by the law of total expectation. 
    
    Our strategy here is to use the high probability bound from Lemma~\ref{lem:concentraion_mc}.
    For some $\delta>N\dmix{K}$ to be determined later, let $\tilde{\delta}=\delta-N\dmix{K}$ and denote the following event:
    \[
        \E = \cbrac{\norm{g_t^N - \nabla F(w_t)}\leq 12G\sqrt{\frac{K}{N}}\brac{1 + \sqrt{\log{(K/\tilde{\delta})}}} + \frac{6GK}{N}}\; .
    \]
    Using the trivial bound $\norm{g_t^{N} - \dF(w_t)}\leq 2G$, by the law of total expectation we have
    \begin{align*}
        \bbE_{t-1}\sbrac{\norm{g_t^N - \nabla F(w_t)}} &= \bbE_{t-1}\sbrac{\norm{g_t^N - \nabla F(w_t)}\mid\E}\cdot\underbrace{\prob_{t-1}\brac{\E}}_{\leq 1} + \bbE_{t-1}[\underbrace{\norm{g_t^N - \nabla F(w_t)}}_{\leq 2G}\mid \E^c]\cdot\prob_{t-1}\brac{\E^c}  \\ &\leq \bbE_{t-1}\sbrac{\norm{g_t^N - \nabla F(w_t)}\mid\E} + 2G\cdot\prob_{t-1}\brac{\E^c} \\ &\leq 12G\sqrt{\frac{K}{N}}\brac{1 + \sqrt{\log{(K/\tilde{\delta})}}} + \frac{6GK}{N} + 2\delta G \; ,
    \end{align*}
    where in the last inequality we used Lemma~\ref{lem:concentraion_mc} to bound the conditional probability of $\E^c$ given $\F_t$, which applies since $N\geq 2K$. Note that we can bound $N\dmix{K}$ above as follows:
    \begin{align*}
        N\dmix{K} = N\dmix{\taumix\ceil{2\log{T}}} \leq N\cdot 2^{-\ceil{2\log{T}}} \leq \frac{N}{T^{2}}\leq \frac{1}{N}\; ,
    \end{align*}
    where we used Eq.~\eqref{eq:dmix} to bound $\dmix{K}$ and the fact that $N\leq T$. Therefore, by choosing $\delta = 2/N>1/N\geq N\dmix{K}$ we have that $\tilde{\delta}\geq 1/N$, and overall we get
    \begin{align*}
        \bbE_{t-1}\sbrac{\norm{g_{t}^{N} - \dF(w_t)}} &\leq 12G\sqrt{\frac{K}{N}}\brac{1 + \sqrt{\log{(KN)}}} + \frac{6GK}{N} + \frac{4G}{N}\; .
    \end{align*}
    Since $K/N\leq 1/2\leq 1$, the $\sqrt{K/N}$ term is the most dominant term; thus, we conclude that
    \[
        \bbE_{t-1}\sbrac{\norm{g_{t}^{N} - \dF(w_t)}}\leq \Ocaltilde\brac{G\sqrt{\frac{K}{N}}}\; .
    \]
    Using similar derivation, and the trivial bound $\norm{g_t^{N} - \dF(w_t)}^2\leq 4G^2$, we get:
    \begin{align*}
        \bbE_{t-1}\sbrac{\norm{g_t^N - \nabla F(w_t)}^2} &\leq \bbE_{t-1}\sbrac{\norm{g_t^N - \nabla F(w_t)}^2\mid\E} + 4\delta G^2 \\ &\leq \brac{12G\sqrt{\frac{K}{N}}\brac{1 + \sqrt{\log{(KN)}}} + \frac{6GK}{N}}^2 + \frac{8G^2}{N} \\ &\leq \frac{576G^2 K}{N}\brac{1 + \log{(KN)}} + \frac{72G^2 K^2}{N^2} + \frac{8G^2}{N} \\ &= \Ocaltilde\brac{\frac{G^2 K}{N}}\; ,
    \end{align*}
    where we used the inequality $(a+b)^2\leq 2a^2+2b^2$ twice. 
\end{proof}

\section{Proof of Theorem~\ref{thm:convex_convergence}}\label{app:proof_thm_convex_case}
In this section, we provide the full proof for the convergence of \acronym in the convex setting. \\ \\
\textit{Proof of Theorem~\ref{thm:convex_convergence}.} \\ \\
By the convexity of $F$, we have $\bbE[F(w_t) - F(w^*)]\leq \bbE[\dF(w_t)^\top(w_t - w^*)]$. Adding and subtracting $g_t$, we can write
\begin{equation}
    \bbE[F(w_t) - F(w^*)] \leq \bbE\sbrac{g_t^\top(w_t-w^*)} + \bbE\sbrac{(\dF(w_t) - g_t)^\top(w_t-w^*)}\; .
\end{equation}
We bound the second term using the result from Lemma~\ref{lem:mlmc_bounds} and Cauchy-Schwarz inequality. That is, for $j_{\max} = \floor{\log{T}}$, we have
\begin{align}\label{eq:bound_inner_prod}
    \bbE\sbrac{(\dF(w_t) - g_t)^\top(w_t-w^*)} &= \bbE\sbrac{\bbE_{t-1}\sbrac{(\dF(w_t) - g_t)^\top(w_t-w^*)}} \nonumber\\ &= \bbE\sbrac{(\dF(w_t) - \bbE_{t-1}[g_t])^\top(w_t - w^*)} \nonumber\\ &= \bbE\sbrac{(\dF(w_t) - g_t^{j_{\max}})^\top(w_t - w^*)}\nonumber\\ &\leq \bbE\sbrac{\lVert\dF(w_t) - g_t^{j_{\max}}\rVert\norm{w_t - w^*}} \nonumber\\ &\leq D\bbE[\lVert\dF(w_t) - g_t^{j_{\max}}\rVert]\; .
\end{align}
Let $K=\taumix\ceil{2\log{T}}$. Since $T/2\leq 2^{j_{\max}}\leq T$, according to Lemma~\ref{lem:expectation_and_second_order_bound} it holds that
\begin{equation}\label{eq:bound_norm_j_max}
    \bbE\sbrac{\lVert g_t^{j_{\max}} - \dF(w_t)\rVert}\leq \Ocaltilde\brac{G\sqrt{\frac{K}{2^{j_{\max}}}}} = \Ocaltilde\brac{G\sqrt{\frac{\taumix}{T}}}, \quad\forall t\in\sbrac{T}\; .
\end{equation}
Thus, we conclude from Eq.~\eqref{eq:bound_inner_prod} that $\bbE\sbrac{(\dF(w_t) - g_t)^\top(w_t-w^*)} \leq\Ocaltilde(GD\sqrt{\taumix/T})$. By Jensen's inequality, we have
\begin{align}\label{eq:jensen_decom}
    \bbE\sbrac{F(\bar{w}_T) - F(w^*)} &\leq \frac{1}{T}\bbE\sbrac{\sum_{t=1}^{T}{\brac{F(w_t) - F(w^*)}}} \nonumber \\ &\leq \frac{1}{T}\bbE\sbrac{\sum_{t=1}^{T}{g_t^\top(w_t - w^*)}} + \frac{1}{T}\bbE\sbrac{\sum_{t=1}^{T}{(\dF(w_t) - g_t)^\top(w_t-w^*)}}\; .
\end{align}
As we have shown, each term is the second sum is bounded by $\Ocaltilde\brac{GD\sqrt{{\taumix/T}}}$, so their average is also bounded by $\Ocaltilde\brac{GD\sqrt{{\taumix/T}}}$. The first term is the expected regret of AdaGrad, which according to Lemma~\ref{lem:adagrad_regret_bound} is bounded by $\bbE[D\sqrt{2\sum_{t=1}^{T}{\norm{g_t}^2}}]$. From Lemma~\ref{lem:mlmc_bounds}, we know that $\bbE[\lVert g_t\rVert^2]\leq\Ocaltilde(G^2\taumix)$; thus, using Jensen's inequality once again, we obtain
\begin{align*}
    \bbE\sbrac{D\sqrt{2\sum_{t=1}^{T}{\norm{g_t}^2}}} &\leq D\sqrt{2\sum_{t=1}^{T}{\bbE[\norm{g_t}^2]}} \leq\Ocaltilde\brac{GD\sqrt{\taumix T}}\; .
\end{align*}
Plugging-in back to Eq.~\eqref{eq:jensen_decom} concludes the proof.

\section{Proof of Theorem~\ref{thm:mag_td_convergence}}\label{app:proof_thm_mag_td}
Our proof of Theorem~\ref{thm:mag_td_convergence} relies on the same ideas used for proving Theorem~\ref{thm:convex_convergence}. \\ \\
\textit{Proof of Theorem~\ref{thm:mag_td_convergence}.} \\ \\
Using Lemma~\ref{lem:bhandari}, we have
\[
    (1-\gamma)\bbE\sbrac{\norm{V_{\theta^*} - V_{\theta_t}}_M^2}\leq \bbE[\bar{g}(\theta_t)^\top(\theta^* - \theta_t)] = \bbE[g_t^\top(\theta^* - \theta_t)] + \bbE[\brac{\bar{g}(\theta_t) - g_t}^\top(\theta^* - \theta_t)]\; .
\]
Summing over $t=1,\ldots,T$, dividing by $T$ and using Jensen's inequality, we get
\begin{align*}
    (1-\gamma)\bbE\sbrac{\norm{V_{\theta^*} - V_{\bar{\theta}_T}}_M^2} &\leq \frac{1}{T}\bbE\sbrac{\sum_{t=1}^{T}{g_t^\top(\theta^* - \theta_t)}} + \frac{1}{T}\sum_{t=1}^{T}{\bbE\sbrac{\brac{\bar{g}(\theta_t) - g_t}^\top(\theta^* - \theta_t)}}\; .
\end{align*}
The first expectation is the regret bound of AdaGrad; thus, using Jensen's inequality we have   
\[
    \bbE\sbrac{\sum_{t=1}^{T}{g_t^\top(\theta^* - \theta_t)}} \leq \bbE\sbrac{2\sqrt{2}R\sqrt{\sum_{t=1}^{T}{\norm{g_t}^2}}} \leq 2\sqrt{2}R\sqrt{\sum_{t=1}^{T}{\bbE[\norm{g_t}^2]}} \leq \Ocaltilde\brac{GR\sqrt{\taumix T}}\; .
\]
Using the same arguments used for deriving Eq.~\eqref{eq:bound_inner_prod} and \eqref{eq:bound_norm_j_max}, we can obtain the following bound:
\[
    \bbE\sbrac{\brac{\bar{g}(\theta_t) - g_t}^\top\brac{\theta^*-\theta_t}} \leq \Ocaltilde\brac{GR\sqrt{\frac{\taumix}{T}}}, \quad\forall t\in\sbrac{T}\; ,
\]
which implies that
\[
    (1-\gamma)\bbE\sbrac{\norm{V_{\theta^*} - V_{\bar{\theta}_T}}_M^2} \leq \Ocaltilde\brac{GR\sqrt{\frac{\taumix}{T}}}\; .
\]
Dividing by $1-\gamma$ finishes the proof. $\qedwhite$

\section{Motivating Example: The Time Reversal of the Winning Streak}\label{app:linear_mix_time_example}
In this section, we consider an example of a Markov chain whose mixing time grows linearly with the number of states. Define the following vector of size $n$, 
\begin{equation}\label{eq:statinary_reverse_winning_streak}
    \mu(i) = \begin{cases}
    1/{2^{i}}, &i=1,\ldots,n-1, \\
    1/{2^{n-1}}, &i=n.
    \end{cases}
\end{equation}
We consider the time reversal of the ``winning streak" chain~\citep{lovasz1998mixing,levin2017markov} with $n$ states, $\cbrac{1,\ldots,n}$, which has the following transition probabilities:
\begin{align*}
    P(1, i) &= \mu(i), \quad i=1,\ldots,n, \\
    P(i, i-1) &= 1, \quad i=2,\ldots,n-1, \\ 
    P(n, n) = P(n, n-1) &= 1/2\; .
\end{align*}
The following arguments follow \citet{levin2017markov}. It is straightforward to show that $\mu$, as defined in Eq.~\eqref{eq:statinary_reverse_winning_streak}, is the stationary distribution of this chain. Moreover, it can be easily shown that after $n-1$ steps, the distribution of the chain is stationary, regardless of the initial distribution. This implies that the mixing time is upper bounded by $n-1$. Note that if the chain starts at state $n$ and leaves it after one step, then after $n-3$ additional steps, it will be at state $2$. Therefore, we have that $P^{n-2}(n, 2) = P(n, n-1)=1/2$, which according to the definition of the total variation distance implies that $\dmix{n-2}\geq\lvert P^{n-2}(n,2) - \mu(2)\rvert = 1/4$. Overall, we conclude that for this chain $\taumix=\Theta(n)$. 

This example demonstrates that the mixing time can grow polynomially with the size of the problem.

\section{Mixing Time of Markov Chains}\label{app:MC_mixing_time}
In this section we state a classical result from the theory of Markov chains and mixing times that gives a lower and upper bounds on the mixing time of a markov chain based on properties of its transition matrix. Then, we use this result to bound the mixing time of the $2$-state chain used in our experiments (Section~\ref{sec:experiments}).

\begin{theorem}[\citealp{levin2017markov}, Theorems 12.4 and 12.5]\label{thm:mixing_time_transition_matrix}
    Consider an ergodic and reversible Markov chain with $n$ states, whose transition matrix is $P$, and let $1=\lambda_1>\lambda_2\geq\ldots\geq\lambda_n$ be the eigenvalues of $P$. Moreover, let $\mu_{\min} = \min_{i\in\cbrac{1,\ldots,n}}{\mu(i)}$ be the minimal stationary distribution of the chain. Then, the mixing time of the chain can be bounded as follows:
    \begin{equation*}
        \frac{\abs{\lambda_2}}{1 - \max\{\abs{\lambda_2}, \abs{\lambda_n}\}}\log{2}\leq\taumix\leq\frac{1}{1 - \max\{\abs{\lambda_2}, \abs{\lambda_n}\}}\log\brac{\frac{4}{\mu_{\min}}}\; .
    \end{equation*}
\end{theorem}

Recall that the transition matrix of the 2-state chain used in our experiments is given by 
\[
    P = \brac{\begin{array}{cc}
        1-p & p \\
        p & 1-p
    \end{array}}\; ,
\]
for some $p\in\brac{0,1}$. Since the chain is symmetric, the stationary distribution is uniform, $\mu=\brac{\frac{1}{2}, \frac{1}{2}}$, and the eigenvalues of this matrix are $\lambda_1=1, \lambda_2=1-2p$. Therefore, the result in Theorem~\ref{thm:mixing_time_transition_matrix} implies that the mixing time of the chain satisfies
\[
    \frac{1-2p}{2p}\log{2}\leq\taumix\leq\frac{1}{2p}\log{8}\; .
\]
Therefore, if $p$ is very small then $1-2p\approx1$ and $\taumix=\Theta(1/p)$. 

\end{document}